\def\year{2020}\relax
\documentclass[letterpaper]{article} % DO NOT CHANGE THIS
\usepackage{aaai20}  % DO NOT CHANGE THIS
\usepackage{times}  % DO NOT CHANGE THIS
\usepackage{helvet} % DO NOT CHANGE THIS
\usepackage{courier}  % DO NOT CHANGE THIS
\usepackage[hyphens]{url}  % DO NOT CHANGE THIS
\usepackage{graphicx} % DO NOT CHANGE THIS
\usepackage{graphicx}  % DO NOT CHANGE THIS
\usepackage{subfigure}
\usepackage{xspace}
\usepackage{multirow}
\usepackage{amsmath}

\usepackage{bm}
\usepackage{soul}
\usepackage[normalem]{ulem}        % Part of the standard distribution
\soulregister\cite7 
\soulregister\citep7 
\soulregister\citet7
\soulregister\ref7 
\soulregister\pageref7

\DeclareMathAlphabet\mathbfcal{OMS}{cmsy}{b}{n}

\newcommand{\beijing}{{\sc Beijing}\xspace}

\newcommand{\shenzhen}{{\sc Shenzhen}\xspace}
\newcommand{\hmgnn}{{\sc SHARE}\xspace}

\newcommand{\eg}{\emph{e.g.},\xspace}
\newcommand{\ie}{\emph{i.e.},\xspace}

\newcommand{\etc}{\emph{etc.}\xspace}

\newcommand\figref[1]{Figure~\ref{#1}}

\newcommand\tabref[1]{Table~\ref{#1}}

\newcommand\equref[1]{Equation~(\ref{#1})}

\newtheorem{pro}{Problem}
\newtheorem{defi}{Definition}

\title{Semi-Supervised Hierarchical Recurrent Graph Neural Network \\
for City-Wide Parking Availability Prediction}
\author{Weijia Zhang\textsuperscript{\rm 1}\thanks{Equal contribution.}, Hao Liu\textsuperscript{\rm 2}\footnotemark[1]\thanks{Corresponding author.}, Yanchi Liu\textsuperscript{\rm 3}, Jingbo Zhou\textsuperscript{\rm 2}, Hui Xiong\textsuperscript{\rm 2}\footnotemark[2]
\\\textsuperscript{\rm 1}University of Science and Technology of China, Hefei, China,  \textsuperscript{\rm 2}Business Intelligence Lab, Baidu Research, \\ National Engineering Laboratory of Deep Learning Technology and Application, Beijing, China,  \textsuperscript{\rm 3}Rutgers University, USA
\\wjzhang3@mail.ustc.edu.cn, \{liuhao30, zhoujingbo\}@baidu.com, yanchi.liu@rutgers.edu, xionghui@gmail.com
}

\begin{document}

\maketitle
\begin{abstract}
The ability to predict city-wide parking availability is crucial for the successful development of Parking Guidance and Information~(PGI) systems. Indeed, the effective prediction of city-wide parking availability can improve parking efficiency, help urban planning, and ultimately alleviate city congestion. However, it is a non-trivial task for predicting city-wide parking availability because of three major challenges: 1) the non-Euclidean spatial autocorrelation among parking lots, 2) the dynamic temporal autocorrelation inside of and between parking lots, and 3) the scarcity of information about real-time parking availability obtained from real-time sensors~(\eg camera, ultrasonic sensor, and GPS).
To this end, we propose \emph{\underline{S}emi-supervised \underline{H}ier\underline{a}rchical \underline{Re}current Graph Neural Network}~(\hmgnn) for predicting city-wide parking availability. Specifically, we first propose a hierarchical graph convolution structure to model non-Euclidean spatial autocorrelation among parking lots.
Along this line, a contextual graph convolution block and a soft clustering graph convolution block are respectively proposed to capture local and global spatial dependencies between parking lots.
Additionally, we adopt a recurrent neural network to incorporate dynamic temporal dependencies of parking lots.
Moreover, we propose a parking availability approximation module to estimate missing real-time parking availabilities from both spatial and temporal domain. 
Finally, experiments on two real-world datasets demonstrate the prediction performance of \hmgnn outperforms seven state-of-the-art baselines.
\end{abstract}

% Introduction
\section{Introduction}\label{sec:intro}
%%%%%%%%%%%%%%%%%%%%%%%%%%%%% Parking is important %%%%%%%%%%%%%%%%%%%%%%%%%%%%%%%%
In recent years, we have witnessed significant development of Intelligent Transportation Systems~(ITS)~\cite{zhang2011data}.
Parking guidance and information~(PGI) systems, especially parking availability prediction, is an indispensable component of ITS.
According to a survey by the International Parking Institute~(IPI)\footnotemark[1], over $30\%$ cars on the road are searching for parking, and these cruising cars contribute up to $40\%$ traffic jams in urban areas~\cite{shoup2006cruising}.
Thus, city-wide parking availability prediction is of great importance to help drivers efficiently find parking, help governments for urban planning, and alleviate the city's traffic congestion.

%%%%%%%%%%%%%%%%% Related work & limitation %%%%%%%%%%%%%%%%%%%%%%%%%%\
Due to its importance, city-wide parking availability prediction has attracted much attention from both academia and industry.
On one hand, Google Maps predicts parking difficulty on a city-wide scale based on users' survey and trajectory data~\cite{arora2019hard}, and Baidu Maps estimates real-time city-wide parking availability based on environmental contextual features~(\eg Point of Interest~(POI), map queries, \etc)~\cite{rong2018parking}.
The above mentions make city-wide parking availability prediction based on biased and indirect input signals~(\eg user's feedback are noisy and lagged), which may induce inaccurate prediction results.
On the other hand, in recent years, we have witnessed real-time sensor devices such as camera, ultrasonic sensor, and GPS become ubiquitous, which can significantly improve the prediction accuracy of parking availability~\cite{mathur2010parknet,fusek2013adaboost,zhou2015smiler}.
However, for economic and privacy concerns, it is difficult to be scaled up to cover all parking lots of a city.

\footnotetext[1]{https://www.parking.org/wp-content/uploads/2015/12/Emergi\\ng-Trends-2012.pdf}

%%%%%%% Parking forecasting is challenging with our setting %%%%%%%%%%
In this paper, we propose to \emph{simultaneously} predict the availability of each parking lot of a city, based on both environmental contextual data~(\eg POI distribution, population) and partially observed real-time parking availability data. By integrating both datasets, we can make a better parking availability prediction at a city-scale.
However, it is a non-trivial task faced with the following three major challenges.
(1) \emph{Spatial autocorrelation}. The availability of a parking lot is not only effected by the occupancy of nearby parking lots but may also synchronize with distant parking lots~\cite{Wang2017kdd,liu2017point}. The first challenge is how to model the irregular and non-Euclidean autocorrelation between parking lots.
(2) \emph{Temporal autocorrelation}. Future availability of a parking lot is correlated with its availability of previous time periods~\cite{rajabioun2015street}. Besides, the spatial autocorrelation between parking lots may also vary over time~\cite{liang2018geoman,yao2019revisiting}. How to model dynamic temporal autocorrelation of each parking lot is another challenge.
(3) \emph{Parking availability scarcity}. Only a small portion of parking lots are equipped with real-time sensors. According to one of the largest map service application, there are over $70,000$ parking lots in Beijing, however, only $6.12\%$ of them have real-time parking availability data.
The third challenge is how to utilize the scarce and incomplete real-time parking availability information.

%%%%%%%%%%%%%%%%%%%%%%% Our approach & contribution %%%%%%%%%%%%%%%%%%%%%%%%%%%%%%%
To tackle above challenges, in this paper, we present \emph{\underline{S}emi-supervised \underline{H}ier\underline{a}rchical \underline{Re}current Graph Neural Network}~(\hmgnn) for city-wide parking availability prediction. Our major contributions are summarized as follows:
\begin{itemize}
\item We propose a semi-supervised spatio-temporal learning framework to incorporate both environmental contextual factors and sparse real-time parking availability data for city-wide parking availability prediction.
\item We propose a hierarchical graph convolution module to capture non-Euclidean spatial correlations among parking lots.
It consists of a contextual graph convolution block and a soft clustering graph convolution block for local and global spatial dependencies modeling, respectively.
\item We propose a parking availability approximation module to estimate missing real-time parking availabilities of parking lots without sensor monitoring. 
Specifically, we introduce a propagating convolution block and reuse the temporal module to approximate missing parking availabilities from both spatial and temporal domain, then fuse them through an entropy-based mechanism.
\item We evaluate \hmgnn on two real-world datasets collected from \beijing and \shenzhen, two metropolises in China. The results demonstrate our model achieves the best prediction performance against seven baselines.
\end{itemize}

% Preliminaries
\section{Preliminaries}\label{sec:preliminary}
Consider a set of parking lots $P=P_l \cup P_u = \{p_1, p_2, \dots, p_N\}$, where $N$ is the total number of parking lots, $P_l$ and $P_u$ denote a set of parking lots with and without real-time sensors~(\eg camera, ultrasonic sensor, GPS, \etc), respectively. Let $\mathbf{X}^t =\{\mathbf{x}^t_1, \mathbf{x}^t_2, \dots, \mathbf{x}^t_N\} \in \mathcal{R}^{N \times M}$ denote observed $M$ dimensional contextual feature vectors~(\eg POI distribution, population, \etc) for all parking lots in $P$ at time $t$. 
We begin the formal definition of parking availability prediction with the definition of parking availability.

\begin{defi}
	\textbf{Parking availability (PA)}. Given a parking lot $p_i\in P$, at time step $t$, the parking availability of $p_i$, denoted $y^t_i$ is defined as the number of vacant parking spot in $p_i$.
\end{defi}

Specifically, we use $\mathbf{y}^t_{P_l}=\{y^t_1, y^t_2,\dots, y^t_{|P_l|}\}$ to denote observed PAs of parking lots in $P_l$ at time step $t$. 
In this paper, we are interested in predicting PAs for all parking lots $p_i\in P$ by leveraging the contextual data of $P$ and partially observed real-time parking availability data of $P_l$.

\begin{pro}
	\textbf{Parking availability prediction problem}. 
	%Given $\mathcal{G}=(\mathcal{V}, \mathcal{E}, \mathbf{A})$,
	Given historical time window $T$, contextual features for all parking lots $\mathbfcal{X}=(\mathbf{X}^{t-T+1}, \mathbf{X}^{t-T+2}, \dots, \mathbf{X}^{t})$, and partially observed real-time PAs $\mathbfcal{Y}_{P_l}=(\mathbf{y}^{t-T+1}_{P_l}, \mathbf{y}^{t-T+2}_{P_l}, \dots, \mathbf{y}^{t}_{P_l})$, our problem is to predict PAs for all $p_i \in P$ over the next $\tau$ time steps,
	\begin{equation}
	f(\mathbfcal{X}; \mathbfcal{Y}_{P_l}) \\
	\rightarrow (\mathbf{\hat{y}}^{t+1}, \mathbf{\hat{y}}^{t+2}, \dots, \mathbf{\hat{y}}^{t+\tau}),
    \end{equation}
where $\mathbf{\hat{y}}^{t+1}=\mathbf{\hat{y}}^{t+1}_{P_l} \cup \mathbf{\hat{y}}^{t+1}_{P_u}$, $f(\cdot)$ is the mapping function we aim to learn. 
\end{pro}

% --Framework
\section{Framework overview}\label{sec:framework}
\begin{figure}[tbp]
  \centering
  \includegraphics[width=1.0\columnwidth]{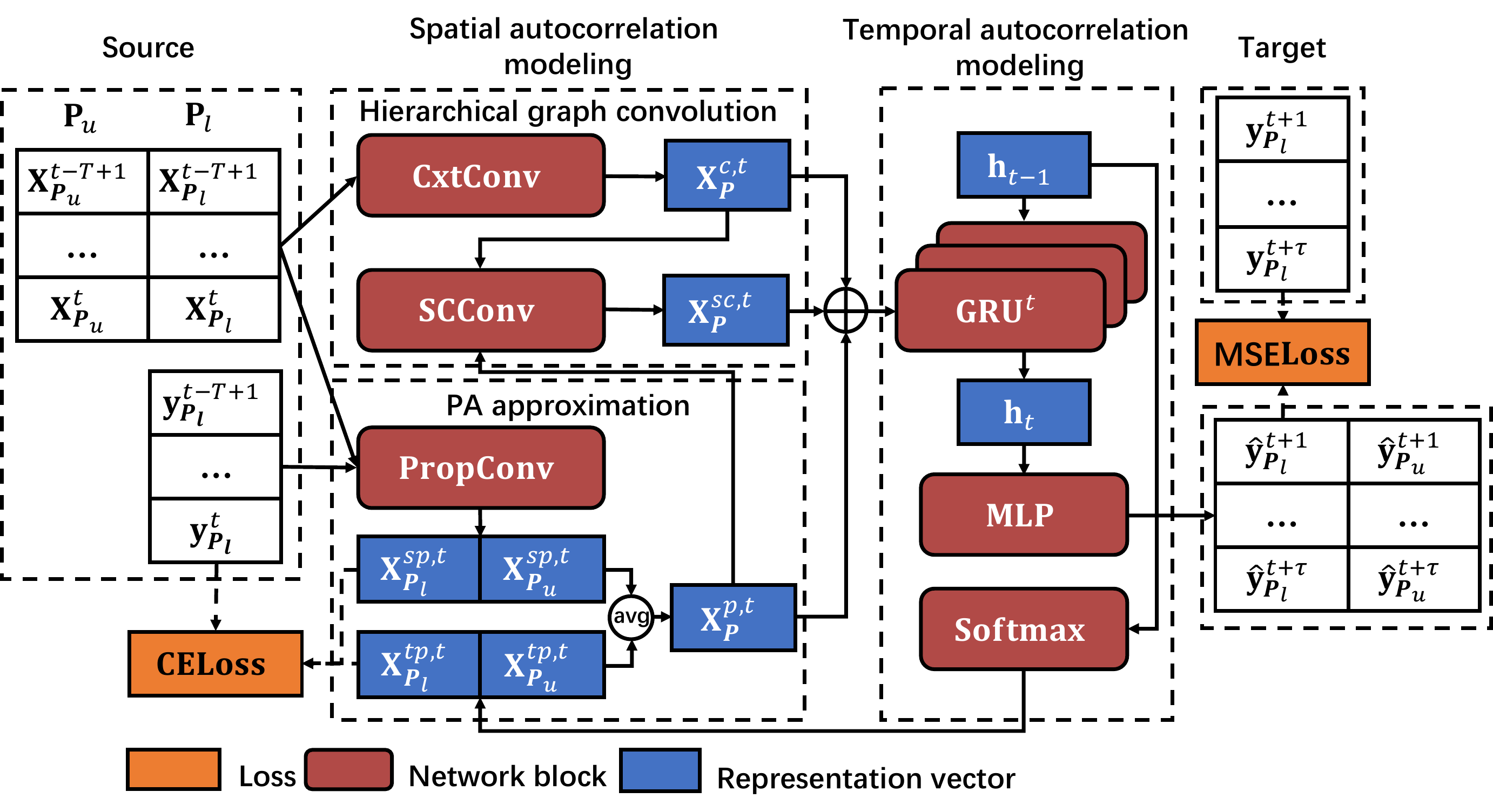}
  \caption{The framework overview of \hmgnn.}
  \label{fig:hmgnn}
\end{figure}

The architecture of \hmgnn is shown in \figref{fig:hmgnn}, where the inputs are contextual features as well as partially observed real-time PAs, and the output are the predicted PAs of all parking lots in next $\tau$ time steps. 
There are three major components in \hmgnn.
First, the \emph{Hierarchical graph convolution} module models spatial autocorrelations among parking lots, where the \emph{Contextual Graph Convolution}~(CxtConv) block captures local spatial dependencies between parking lots through rich contextual features~(\eg POI distribution, regional population, \etc), while the \emph{Soft Clustering Graph Convolution}~(SCConv) block captures global correlations among distant parking lots by softly assigning each parking lot to a set of latent cluster nodes.
Second, the temporal autocorrelation modeling module employs the \emph{Gated Recurrent Unit~(GRU)} to model dynamic temporal dependencies of each parking lot. 
Third, the \emph{PA approximation} module estimates distributions of missing PAs for parking lots in $P_u$, from both spatial and temporal domain.
In the spatial domain, the \emph{Propagating Graph Convolution}~(PropConv) block propagates observed real-time PAs to approxinate missing PAs based on the contextual similarity of each parking lot. In the temporal domain, we reuse the GRU module to approximate current PA distributions based on its output in previous time period. Two estimated PA distributions are then fused through an entropy-based mechanism and feed to SCConv block and GRU module for final prediction.

% -- Model
\section{Hierarchical spatial dependency modeling}\label{sec:spatial}
We first introduce the hierarchical graph convolution module, including the contextual graph convolution block and the soft clustering graph convolution block.

\subsection{Contextual graph convolution}
In the spatial domain, the PA of nearby parking lots are usually correlated and mutually influenced by each other. 
For example, when there is a big concert, the PAs of parking lots near the concert hall are usually low, and the parking demand usually gradually diffuses from nearby to distant.
Inspired by the recent success of graph convolution network~\cite{kipf2017semi,velivckovic2017graph} on processing non-Euclidean graph structures, we first introduce the CxtConv block to capture local spatial dependencies solely based on contextual features.

We model the local correlations among parking lots as a graph $G=(V, E, A)$, where $V=P$ is the set of parking lots, $E$ is a set of edges indicating connectivity among parking lots, and $A$ denotes the proximity matrix of $G$ \cite{ma2019efficient}. Specifically, we define the connectivity constraint $e_{ij}\in E$ as
\begin{equation}\label{equ:cxtedge}
e_{ij}=\left\{
\begin{aligned}
    &1,\quad dist(v_{i},v_{j}) \leq \epsilon\\
    &0,\quad otherwise
\end{aligned},
\right.
\end{equation}
where $dist(\cdot)$ is the road network distance between parking lots $p_i$ and $p_j$, $\epsilon$ is a distance threshold. 

Since the influence of different nearby parking lots may vary non-linearly, we employ an attention mechanism to compute the coefficient between parking lots, defined as
\begin{equation}
	c_{ij} = Attn(\mathbf{W}_{a}\mathbf{x}^{c}_i, \mathbf{W}_{a}\mathbf{x}^{c}_j),
\end{equation}
where $\mathbf{x}^{c}_i$ and $\mathbf{x}^{c}_j$ are current contextual representations of parking lot $p_i$ and $p_j$, $\mathbf{W}_{a}$ is a learnable weighted matrix shared over all edges, and $Attn(\cdot)$ is a shared attention mechanism~(\eg dot-product, concatenation,  \etc)~\cite{Vaswani:2017:AYN:3295222.3295349}.
The proximity score between $p_i$ and $p_j$ is further defined as 
\begin{equation}\label{equ:attention}
	\alpha_{ij} = \frac{exp(c_{ij})}{\sum_{k\in \mathcal{N}_i}exp(c_{ik})}.
\end{equation}

In general, the above attention mechanism is capable of computing pair-wise proximity score for all $p_i\in P$. However, this formulation will lead to quadratic complexity. To weigh more attention on neighboring parking lots and help faster convergence, we inject the adjacency constraint where
the attention operation only operate on adjacent nodes $j\in \mathcal{N}_i$, where $\mathcal{N}_i$ is a set of neighboring parking lots of $p_i$ in $G$.
Note that the influence of nearby parking lot at different time step may also vary, we learn a different proximity score for each different time steps.

Once $\alpha_{ij}$ is obtained, the contextual graph convolution operation updates representation of current parking lot by aggregating and transforming its neighbors, defined as
\begin{equation}\label{equ:cxtconv}
	\mathbf{x}_i^{c'} = \sigma (\sum_{j\in \mathcal{N}_i} \alpha_{ij} \mathbf{W}_{c} \mathbf{x}^{c}_{j}),
\end{equation}
where $\sigma$ is a non-linear activation function, and $\mathbf{W}_{c}\in \mathcal{R}^{d \times d}$ is a learnable weighted matrix shared over all parking lots. 
Note that we can stack $l$ identical contextual graph convolution layers to capture $l$-hop local dependencies, and $\mathbf{x}^{c}_{j}$ is the raw contextual feature in the first CxtConv layer.

\subsection{Soft clustering graph convolution}\label{sec:scconv}
\begin{figure}[t]
  \centering
  \includegraphics[width=0.9\columnwidth]{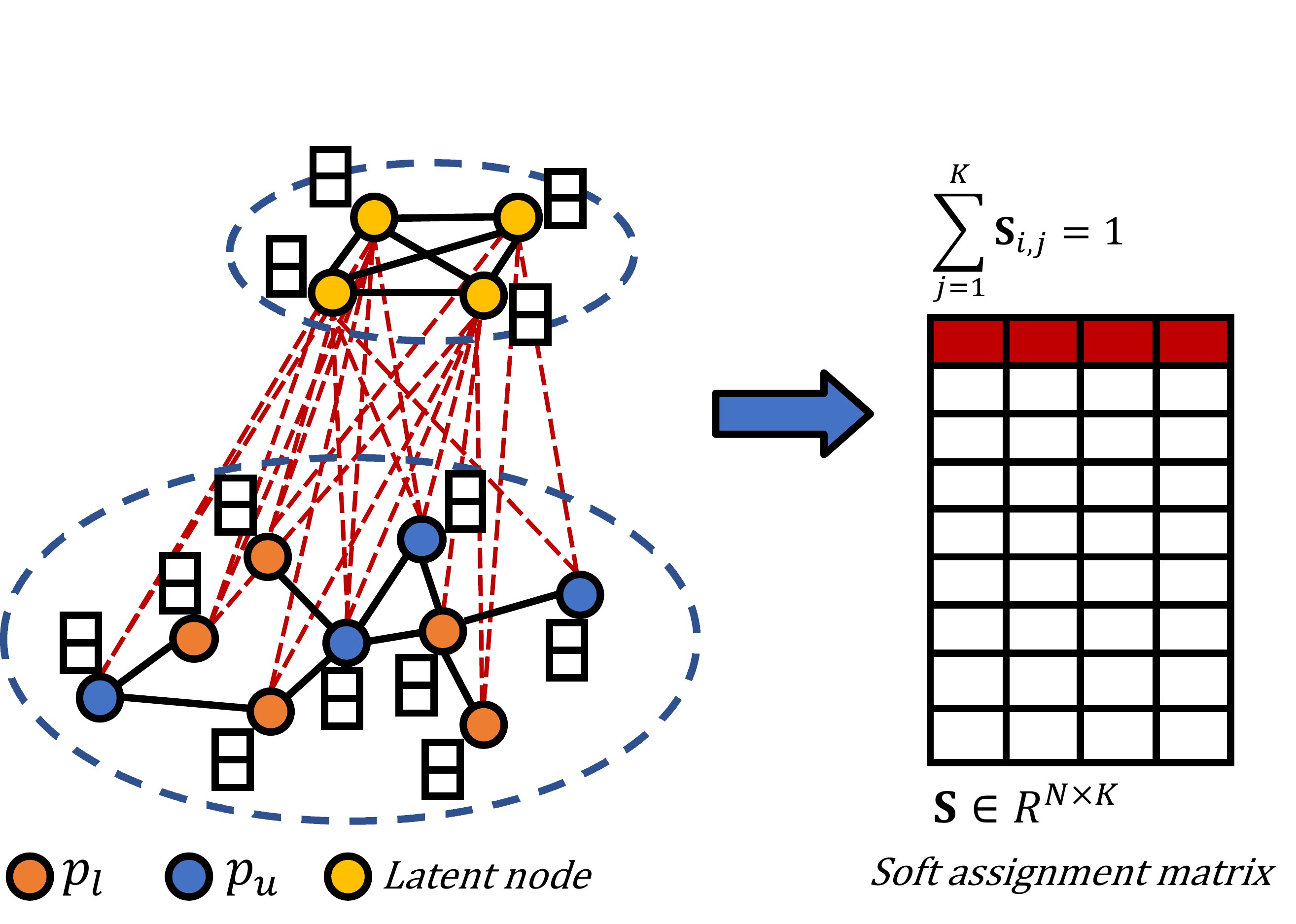}
  \caption{Hierarchical soft clustering.}
  \label{fig:hierarchical}
\end{figure}

Besides local correlation, distant parking lots may also be correlated. For example, distant parking lots in similar functional areas may show similar PA, \eg business areas may have lower PA at office hour, and residential areas may have higher PA at the same time.
However, CxtConv only captures local spatial correlation.
\cite{li2018deeper} shows when $l$ goes large, the representation of all parking lots tends to be similar, therefore losses discriminative power.
To this end, we propose the SCConv block to capture global correlations between parking lots. 
Specifically, SCConv defines a set of latent nodes and learns the representation of each latent node based on learned representations of each parking lot.
Rather than cluster each parking lot into a specific cluster, we learn a soft assignment matrix so that each parking lot have a chance to belong to multiple clusters with different probabilities~(but with total probability equal to one), as shown in \figref{fig:hierarchical}. 

The intuition behind SCConv is two-fold.
First, distant parking lots may have similar contextual features and PAs, therefore should have similar representations. The shared latent node representation can be viewed as a regularization for the prediction task.
Second, one parking lot may be mapped to multiple latent nodes. If we view each latent node as a different functionality class, a parking lot may serve for several functionalities. For example, a parking lot in a recreational center may be occupied by external visitors from a nearby office building.

The key component in SCConv is the soft assignment matrix.
Given that there are $K$ latent nodes, let $\mathbf{S} \in \mathcal{R}^{N \times K}$ denotes the soft assignment matrix, where $\mathbf{S}_{i,j} \in \mathbf{S}$ denotes the probability of $i$-th parking lot $p_i$ maps to $j$-th latent node. Specifically, we use $\mathbf{S}_{i,\cdot}$ denote the $i$-th row and $\mathbf{S}_{\cdot,j}$ denote the $j$-th column of $\mathbf{S}$.
Given the learned representation of each parking lot $\mathbf{x}_i$, each row of $\mathbf{S}$ is computed as
\begin{equation}
	\mathbf{S}_{i,\cdot}=Softmax(\mathbf{W}_s\mathbf{x}_i),
\end{equation}
which guarantees that the probabilities that a given parking lot belongs to each latent node sum equals one. 

%How to conv
Once $\mathbf{S}$ is obtained, the representation of each latent node $\mathbf{x}^s_i \in \mathbf{X}^s$ can be derived by
\begin{equation}
	\mathbf{x}^s_i= \sum_{j=1}^N \mathbf{S}_{i,j}^{\top} \mathbf{x}_j.
\end{equation}

Given the representation of each latent node, similar to CxtConv, we apply soft clustering convolution operation to capture the dependency between each latent node,
\begin{equation}\label{equ:scconv}
	\mathbf{x}_i^{s'} = \sigma(\sum_{j\in \mathcal{N}_i}\alpha^s_{ij}\mathbf{W}_l\mathbf{x}^{s}_j),
\end{equation}
where $\sigma$ is non-linear activation function, and $\alpha^s_{ij}$ is the proximity score between two latent nodes. Rather than introduce extra attention parameter as in CxtConv, we derive proximity score between latent nodes based on adjacency constraint between parking lots,
\begin{equation}
	\alpha^s_{ij}=\sum_{m=1}^N \sum_{n=1}^N \mathbf{S}_{i,m}^{\top} a_{mn} \mathbf{S}_{n,j}.
\end{equation}
where $a_{mn}$ equals one if parking lots $p_m$ and $p_n$ are connected.
With learned latent node representation, we generate the soft clustering representation for each parking lot as a reverse process of latent node representation generation,
\begin{equation}
	\mathbf{x}^{sc}_i = \sum_{j=1}^K \mathbf{S}_{i,j} \mathbf{x}^{s'}_j.
\end{equation}

\section{Temporal dependency modeling}\label{sec:temporal}
We leverage the Gated Recurrent Unit~(GRU)~\cite{chung2014empirical}, a simple yet effective variant of recurrent neural network~(RNN), to model the temporal dependency.
Consider previous $T$ step inputs of parking lot $p_i$, $(\mathbf{x}^{t-T+1}_i, \mathbf{x}^{t-T+2}_i, \cdots, \mathbf{x}^{t}_i)$,
we denote the status of $p_i$ at time step $t-1$ and $t$ as $\mathbf{h}^{t-1}_i$ and $\mathbf{h}^{t}_i$, respectively.
The temporal dependency between $\mathbf{h}^{t-1}_i$ and $\mathbf{h}^{t}_i$ can be modeled by
\begin{equation}
\mathbf{h}^{t}_i = (1-\mathbf{z}^{t}_i)\circ\mathbf{h}^{t-1}_i+\mathbf{z}^{t}_i\circ\widetilde{\mathbf{h}}^{t}_i, 
\end{equation}
where $\mathbf{z}^t_i$, $\widetilde{\mathbf{h}}^{t}_i$ are defined as 
\begin{equation} 
\left\{
\begin{aligned}
    &\mathbf{r}^{t}_i = \sigma{(\mathbf{W}_r[\mathbf{h}^{t-1}_i\oplus\mathbf{x}^{t}_i]+\mathbf{b}_r)}\\
    &\mathbf{z}^{t}_i = \sigma(\mathbf{W}_z[\mathbf{h}^{t-1}_i\oplus\mathbf{x}^{t}_i]+\mathbf{b}_z)\\
    &\widetilde{\mathbf{h}}^{t}_i = \tanh(\mathbf{W}_{\widetilde{h}}[\mathbf{r}^{t}_i\circ\mathbf{h}^{t-1}_i\oplus\mathbf{x}^t_i]+\mathbf{b}_{\widetilde{h}})
\end{aligned},
\right.
\label{eq:gru}
\end{equation}
where $\boldsymbol{W}_r$, $\boldsymbol{W}_z$, $\boldsymbol{W}_{\widetilde{h}}$, $\boldsymbol{b}_r$, $\boldsymbol{b}_z$, $\boldsymbol{b}_{\widetilde{h}}$ are learnable parameters, $\oplus$ is the concatenation operation, and $\circ$ denotes Hadamard product. 
Then the hidden state $\mathbf{h}^t_i$ is directly used to predict PAs of next $\tau$ time steps,
\begin{equation}
(\hat{y}_{i}^{t+1}, \hat{y}_{i}^{t+2}, \dots, \hat{y}_{i}^{t+\tau}) = \sigma (\mathbf{W}_o\mathbf{h}^t_i),
\end{equation}
where $\mathbf{W}_o\in\mathcal{R}^{|\mathbf{h}^t_i| \times \tau}$. 

\section{Parking availability approximation}\label{sec:paapprox}
The real-time PA is a strong signal for future PA prediction. However, only a small portion~(\eg $6.12\%$ in Beijing) of real-time PAs can be obtained through real-time sensors, which prevents us directly apply real-time PA as a part of input feature.
To leverage the information hidden in partially observed real-time PA, we approximate missing PAs from both spatial and temporal domain.
The proposed method consists of three blocks, \ie the spatial PropConv block, the temporal GRU block, and the fusion block.
Note that rather than approximate a scalar PA $\hat{y}$, we learn the distribution of PA, $\mathbf{x}^p=P(\hat{y})$, for better information preservation.
Given a PA $y$, we discretize its distribution to a $p$ dimensional one hot vector $\mathbf{y}\in \mathcal{R}^p$.
The objective of the PA approximation is to minimize the difference between $\mathbf{y}$ and $\mathbf{x}^p$.

\subsection{Spatial based PA approximation}
Similar to CxtConv, for each $p_i\in P_u$, the PropConv operation is defined as
\begin{equation}
	\mathbf{x}^{sp}_i = \sum_{j\in \mathcal{N}_i} \alpha_{ij} \mathbf{y}_{j},
\end{equation}
where $\mathbf{x}^{sp}_i$ is the obtained PA distribution, $\alpha_{ij}$ is the proximity score between $p_i$ and $p_j$.
Different from CxtConv, the estimated PA is only aggregated from nearby parking lots with real-time PA, and we preserve the aggregated vector representation without extra activation function. 
The proximity score is computed through same attention mechanism in \equref{equ:attention}, but with a relaxed connectivity constraint 
\begin{equation}\label{equ:diffedge}
e_{ij}=\left\{
\begin{aligned}
    &1,\quad dist(v_{i},v_{j}) \leq max(\epsilon, dist_{knn}(v_i)),i\neq j\\
    &0,\quad otherwise
\end{aligned},
\right.
\end{equation}
where $dist_{knn}(v_i)$ denotes the road network distance between parking lot $p_i$ and its $k$-th nearest parking lot $p_j \in P_l$.
The relaxed adjacency constraint improves node connectivity for more sufficient propagation of observed PA, and therefore alleviates the data scarcity problem.

\subsection{Temporal based PA approximation}
We reuse the output of the GRU block to approximate real-time PA from the temporal domain.
The difference between current PA approximation and future PA prediction is here we employ a different $Softmax$ function.
Remember that in previous step, we have obtained hidden state $\mathbf{h}^{t-1}_i$ from GRU, we directly approximate distribution of PA at $t$ by
\begin{equation}
	\mathbf{x}^{tp,t}_i = Softmax(\mathbf{W}_{tp}\mathbf{h}_i^{t-1}).
\end{equation}
This step doesn't introduce extra computation for GRU, and the \emph{Softmax} layer normalizes $\mathbf{x}^{tp,t}_i$ sum equals one.

\subsection{Approximated PA fusion}
Rather than directly averaging $\mathbf{x}^{sp}_i$ and $\mathbf{x}^{tp}_i$, we propose an entropy-based mechanism to fuse two PA distributions. Specifically, we weigh more on the approximation less uncertainty~\cite{hsieh2015inferring}, \ie the one with smaller entropy. 
Given an estimated PA distribution $\mathbf{x}_i$, its entropy is 
\begin{equation}
H(\mathbf{x}_i) = -\sum_{j=1}^{p} \mathbf{x}_{i}{(j)} \log \mathbf{x}_{i}{(j)},
\label{eq:entropy}
\end{equation}
where $\mathbf{x}_i{(j)}$ represents the $j$-th dimension of $\mathbf{x}_i$.
We fuse two PA distributions $\mathbf{x}^{sp}_i$ and $\mathbf{x}^{tp}_i$ as follow:
\begin{equation}
\mathbf{x}^p_{i} = \frac{exp(-H(\mathbf{x}^{sp}_{i}))\mathbf{x}_{i}^{sp} + exp(-H(\mathbf{x}^{tp}_{i}))\mathbf{x}^{tp}_{i}}
{\mathbf{Z}_i},
\label{eq:temp4}
\end{equation}
where $\mathbf{Z}_i=exp(-H(\mathbf{x}^{sp}_{i}))+exp(-H(\mathbf{x}^{tp}_{i}))$.

The approximated PA distribution $\mathbf{x}^p_i$ is applied for two tasks. First, it is concatenated with the learned representation of the CxtConv and fed to the SCConv block for latent node representation learning.
Second, it is combined with the output of the CxtConv and SCConv, $\mathbf{x}^t_i = \mathbf{x}^{c,t}_i \oplus \mathbf{x}^{sc,t}_i \oplus \mathbf{x}^{p,t}_i$.
We use $\mathbf{x}^t_i$ as the overall representation for each parking lot $p_i \in P$ at time step $t$, and feed it into the GRU module to generate final PA prediction results.

\section{Model training}\label{sec:train}
Since only parking lots $P_l$ are with observed labels, following the semi-supervised learning paradigm,
\hmgnn aims to minimize the \emph{mean square error}~(MSE) between the predicted PA and the observed PA
\begin{equation}
O_1 = \frac{1}{{\tau |P_l|}}\sum_{i=1}^{|P_l|}\sum_{j=1}^{\tau}(\hat{y}^{t+j}_{i} - y_{i}^{t+j})^2.
\label{eq:loss1}
\end{equation}

Additionally, in PA approximation, we introduce extra cross entropy~(CE) loss to minimize the error between the observed PA and approximated PA distributions
~(\ie the spatial and temporal based PA distribution approximation $\mathbf{x}^{sp,t}_i$ and $\mathbf{x}^{tp,t}_i$) 
in current time step $t$,
\begin{equation}
O_2 = -\frac{1}{|P_l|} \sum_{i=1}^{|P_l|} \mathbf{y}^{t}_i \log{\mathbf{x}^{sp,t}_i},
\label{eq:loss2}
\end{equation}

\begin{equation}
O_3 = -\frac{1}{|P_l|} \sum_{i=1}^{|P_l|} \mathbf{y}^{t}_i \log{\mathbf{x}^{tp,t}_i}.
\label{eq:loss3}
\end{equation}

By considering both MSE loss and CE loss, \hmgnn aims to jointly minimize the following objective
\begin{equation}
O = O_1+\beta(O_2 + O_3),
\label{eq:lossall}
\end{equation}
where $\beta$ is the hyper-parameter controls the importance of two CE losses.

% -- Experiments
\section{Experiments}\label{sec:exp}
\subsection{Experimental setup}
\subsubsection{Data description.} 
\begin{figure}[tbp]
  \includegraphics[width=1.0\columnwidth]{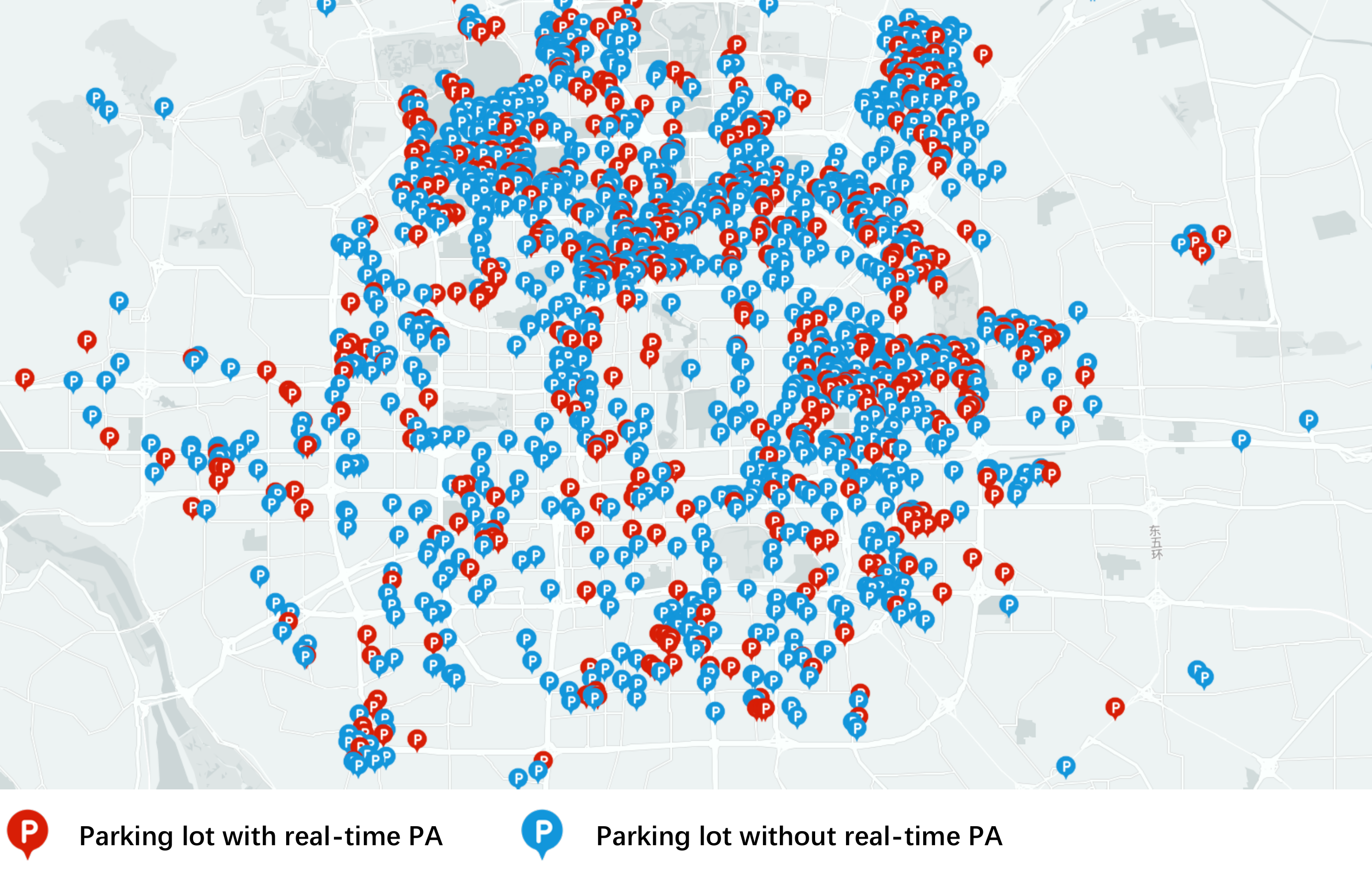}
  \caption{Spatial distribution of parking lots in \beijing.}
  \label{fig:parkinglot_dist}
\end{figure}

We use two real-world datasets collected from \beijing and \shenzhen, two metropolises in China. %, to evaluate \hmgnn. 
Both datasets are ranged from April 20, 2019, to May 20, 2019. All PA records are  crawled every 15 minutes from a publicly accessible app,
%\footnote{http://www.51park.cn/}
in which all parking occupancy information are collected by real-time sensors.
We associate POI distribution~\cite{Hydra,zhu2016days} to each parking lot and aggregate check-in records nearby each parking lot in every $15$ minutes as the population data.
POI and check-in data are collected through Baidu Maps Place API and location SDK \cite{hao2019trans2vec}.
We chronologically order the above data,  take the first $60\%$ as the training set, the following $20\%$ for validation, and the rest as the test set.
In each dataset, $70\%$ parking lots are masked as unlabeled.
The spatial distribution of parking lots in \beijing are shown in~\figref{fig:parkinglot_dist}. The statistics of the datasets are summarized in \tabref{table:dataset}. 

\subsubsection{Implementation details.}
Our model and all seven baselines are implemented with PaddlePaddle.
Following previous work~\cite{li2018dcrnn_traffic,yu2018spatio}, the PA is normalized before input and scaled back to absolute PA in output.
We choose $T=12$ and select $\tau=3$ for prediction.
We set $\epsilon=1$Km and $k=10$ to connect parking lots.
The dimension of $\mathbf{x}^{c}$ and $\mathbf{x}^{sc}$ are fixed to $32$, $p$ is fixed to $50$.
The layer of CxtConv, SCConv, and PropConv are $2,1,1$, respectively. We use dot-product attention in this paper. 
In SCConv, the number of latent nodes is set to $K=0.1N$, where $N$ is the total number of parking lots.
The activation function in CxtConv and SCConv are LeakyReLU~($\alpha=0.2$), and Sigmoid in other layers.
We employ the Adam optimizer for training, fix the learning rate to $0.001$ and set $\beta$ to $0.5$.
For a fair comparison, all parameters of each baseline are carefully tuned based on the recommended settings.

\begin{table}[t]
\small
\centering
\caption{Statistics of datasets.}
\begin{tabular}{c|c|c}
\hline
\textbf{Description} & BEIJING & SHENZHEN\\
\hline
\hline
\# of parking lots & 1,965 & 1,360\\
\hline
\# of PA records & 5,847,840 & 4,047,360\\
\hline
Average \# of parking spots & 210.24 & 185.36 \\
\hline
\# of check-ins & 9,436,362,579 & 3,680,063,509 \\
\hline
\# of POIs & 669,058 & 250,275\\
\hline
\# of POI categories & 197 & 188\\
\hline
\end{tabular}
\label{table:dataset}
\end{table}

\subsubsection{Evaluation metrics.}
We adopt \emph{Mean Average Error}~(MAE) and \emph{Rooted Mean Square Error}~(RMSE), two widely used metrics~\cite{liang2018geoman} for evaluation.

\begin{table*}[t]
\centering
\caption{Parking availability prediction error given by \emph{MAE} and \emph{RMSE} on \beijing and \shenzhen.}
\begin{tabular}{c||c|c||c|c}
\hline
\multirow{2}{*}{\textbf{Algorithm}} & \multicolumn{2}{c||}{\beijing (15/ 30/ 45 min) }&\multicolumn{2}{|c}{\shenzhen (15/ 30/ 45 min)} \\
\cline{2-5}
& MAE & RMSE & MAE & RMSE \\
\hline
\hline
LR & 29.90 / 30.27 / 30.58 & 69.74 / 70.95 / 72.00 
&24.59 / 24.80 / 25.09 & 51.31 / 52.36 / 52.80 \\
\hline
GBRT & 17.29 / 17.81 / 18.40 & 44.60 / 48.50 / 51.59
&13.90 / 14.67 / 14.71 & 35.05 / 37.98 / 38.09 \\
\hline
GRU & 18.51 / 18.78 / 19.73 & 55.43 / 55.92 / 58.64 
& 16.73 / 16.88 / 17.14 & 46.92 / 47.26 / 47.56 \\
\hline
Google-Parking & 21.49 / 21.68 / 22.85 & 57.26 / 59.25 / 60.48 & 17.10 / 18.33 / 18.69 & 47.30 / 48.45 / 49.34 \\
\hline
Du-Parking & 17.67 / 17.70 / 18.03 & 50.17 / 50.63 / 51.75 & 13.91 / 14.17 / 14.39 & 42.66 / 43.24 / 43.56 \\
\hline
STGCN & 16.57 / 16.44 / 17.10 & 50.79 / 51.04 / 52.61 
& 13.46 / 13.59 / 13.88 & 39.26 / 39.96 / 40.29 \\
\hline
DCRNN & 15.66 / 15.97 / 16.30 & 46.28 / 47.80 / 48.87 & 13.11 / 13.19 / 13.89 & 42.74 / 43.37 / 44.27 \\
\hline
CxtGNN~(ours) & 15.29 / 15.69 / 16.15 & 45.55 / 46.69 / 47.78 & 12.39 / 12.73 / 13.09 & 36.31 / 36.92 / 37.46 \\
\hline
CAGNN~(ours) & 12.45 / 12.77 / 13.20 & 39.99 / 40.81 / 41.31 & 10.50 / 10.62 / 10.98 & 31.86 / 32.12 / 32.83 \\
\hline
\textbf{SHARE~(ours)} & \textbf{10.68 / 10.97 / 11.43} & \textbf{32.00 / 32.78 / 33.78} &
\textbf{9.23 / 9.41 / 9.66} & 
\textbf{30.44 / 30.90 / 31.70}\\
\hline
\end{tabular}
\label{table:overall}
\end{table*}

\begin{figure*}[t]
\centering
\subfigure[{\small Ratio of labeled parking lot}]{\label{exp:para-lr_bj}
\includegraphics[width=0.5\columnwidth]{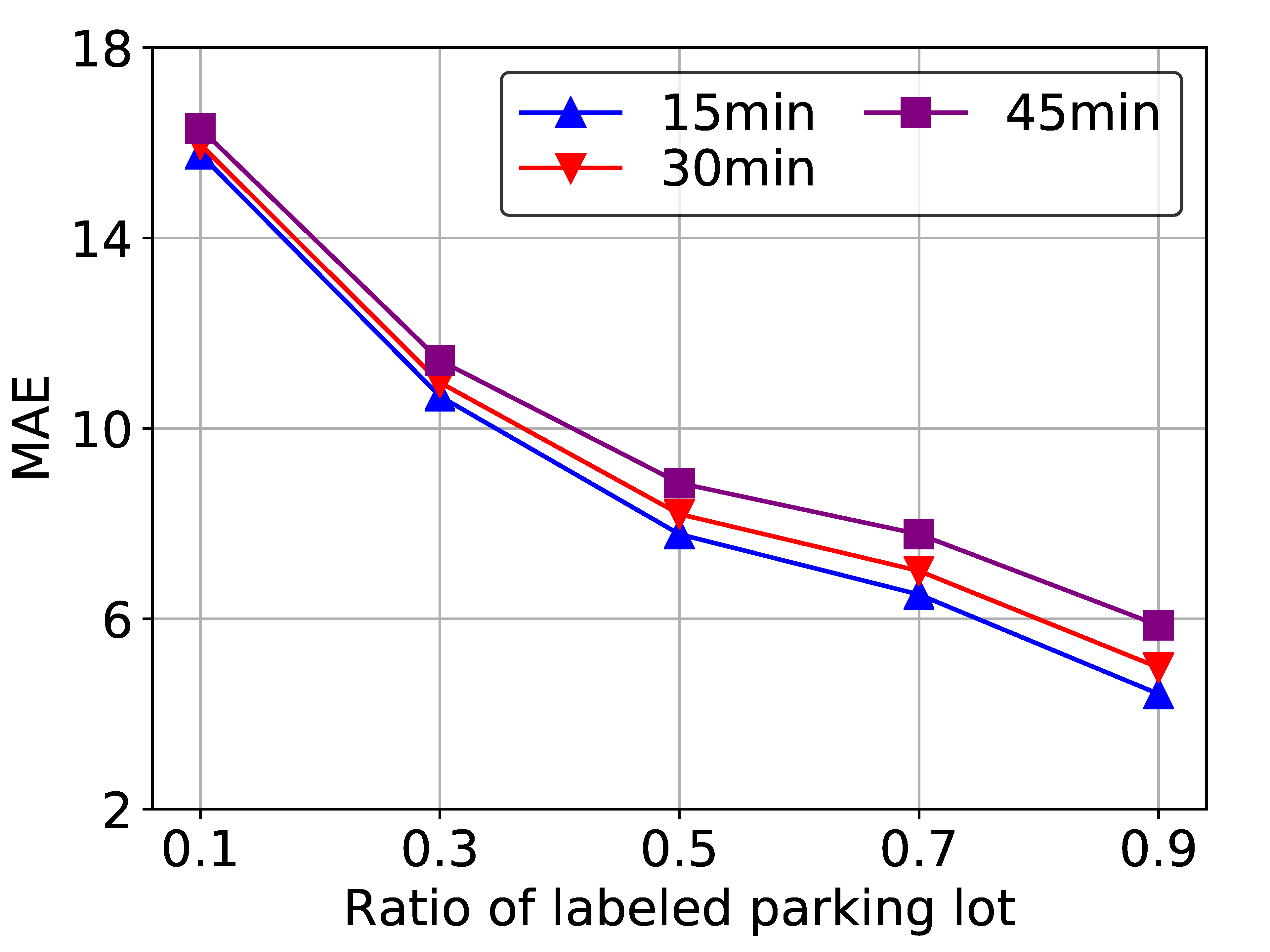}}
\subfigure[{\small Ratio of latent node}]{\label{exp:para-hc-bj}
\includegraphics[width=0.5\columnwidth]{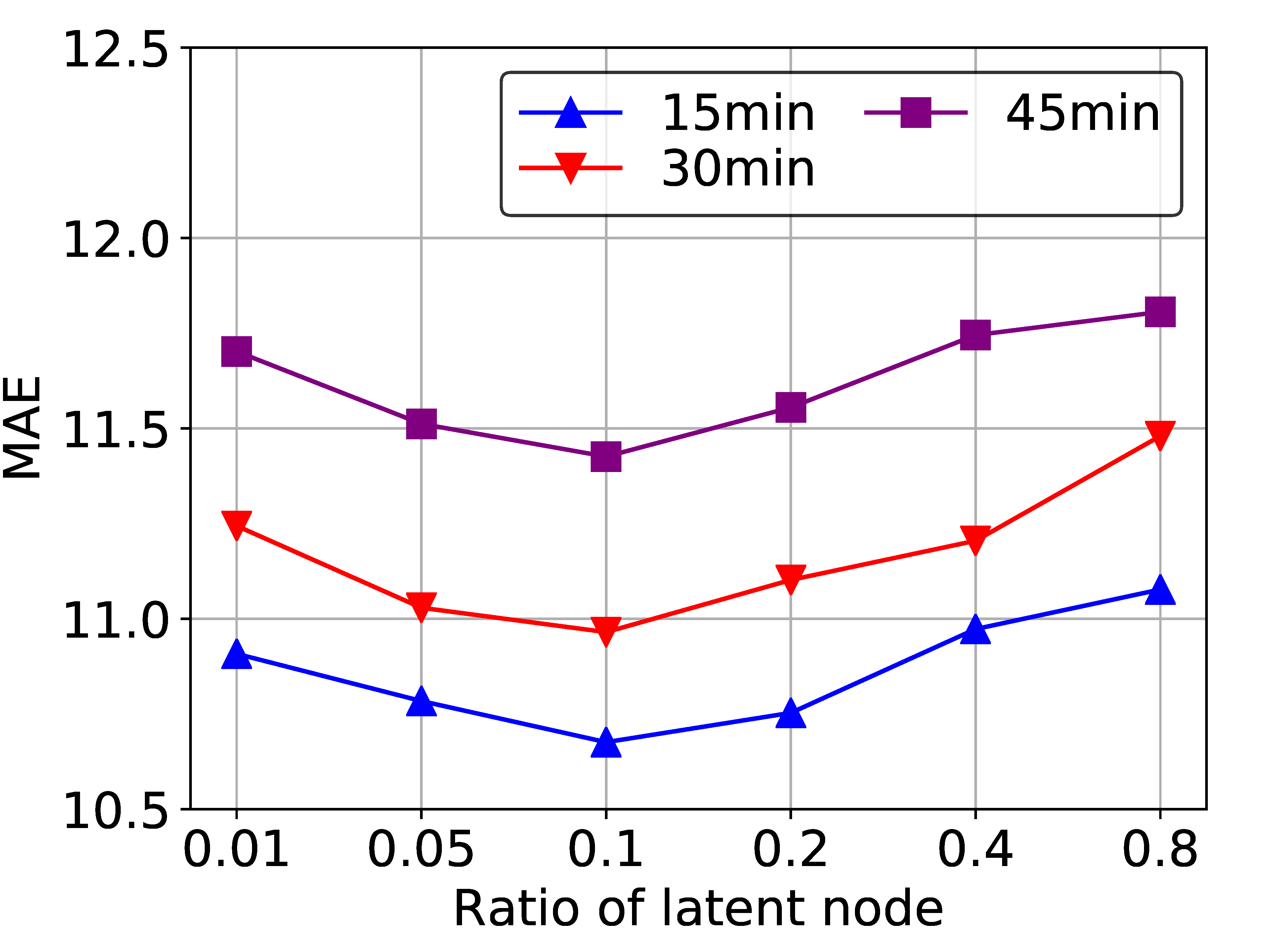}}
\subfigure[{\small Effect of $T$}]{\label{exp:para-tin-bj}
\includegraphics[width=0.5\columnwidth]{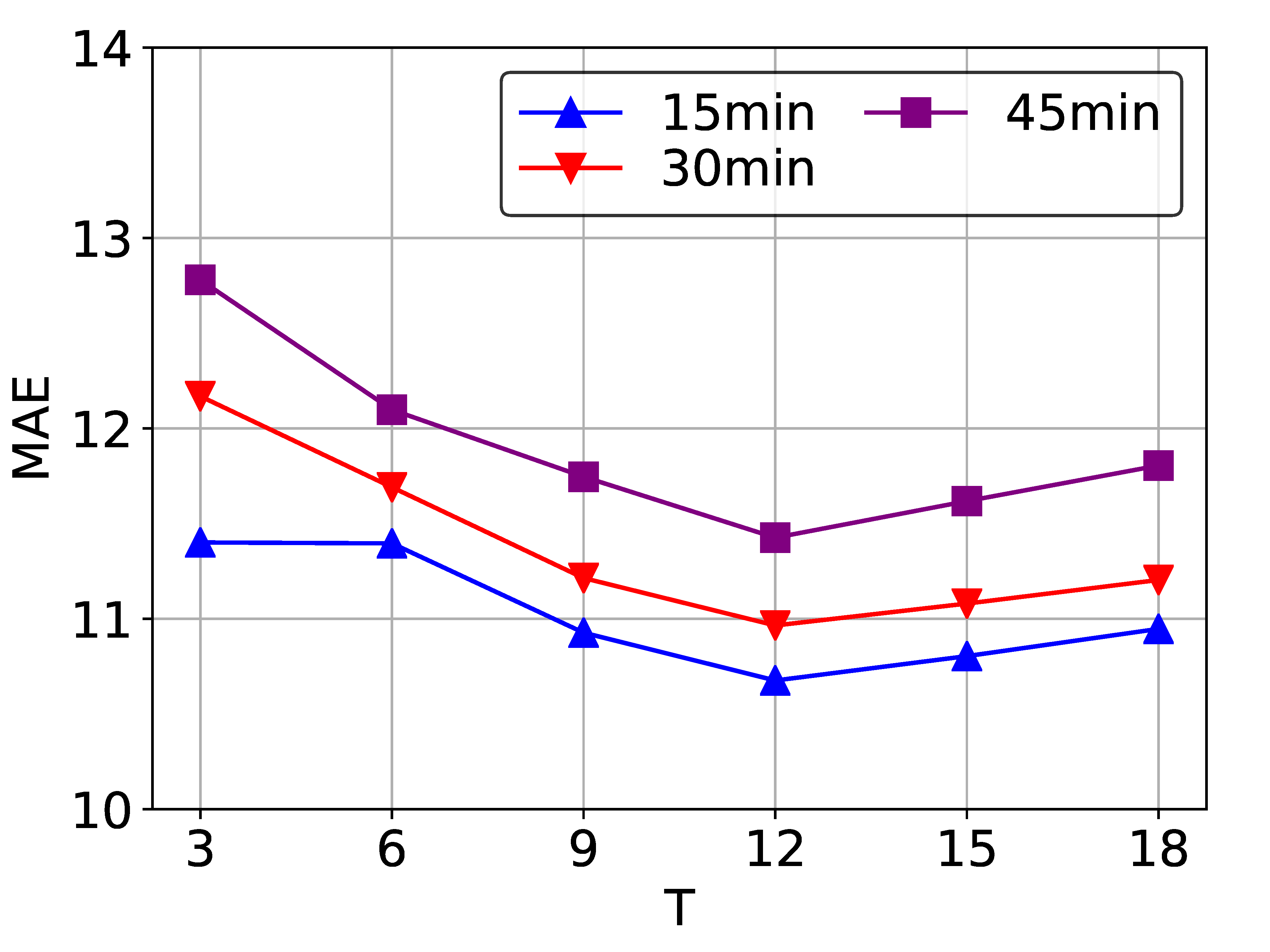}}
\subfigure[{\small Effect of $\tau$}]{\label{exp:para-tout-bj}
\includegraphics[width=0.5\columnwidth]{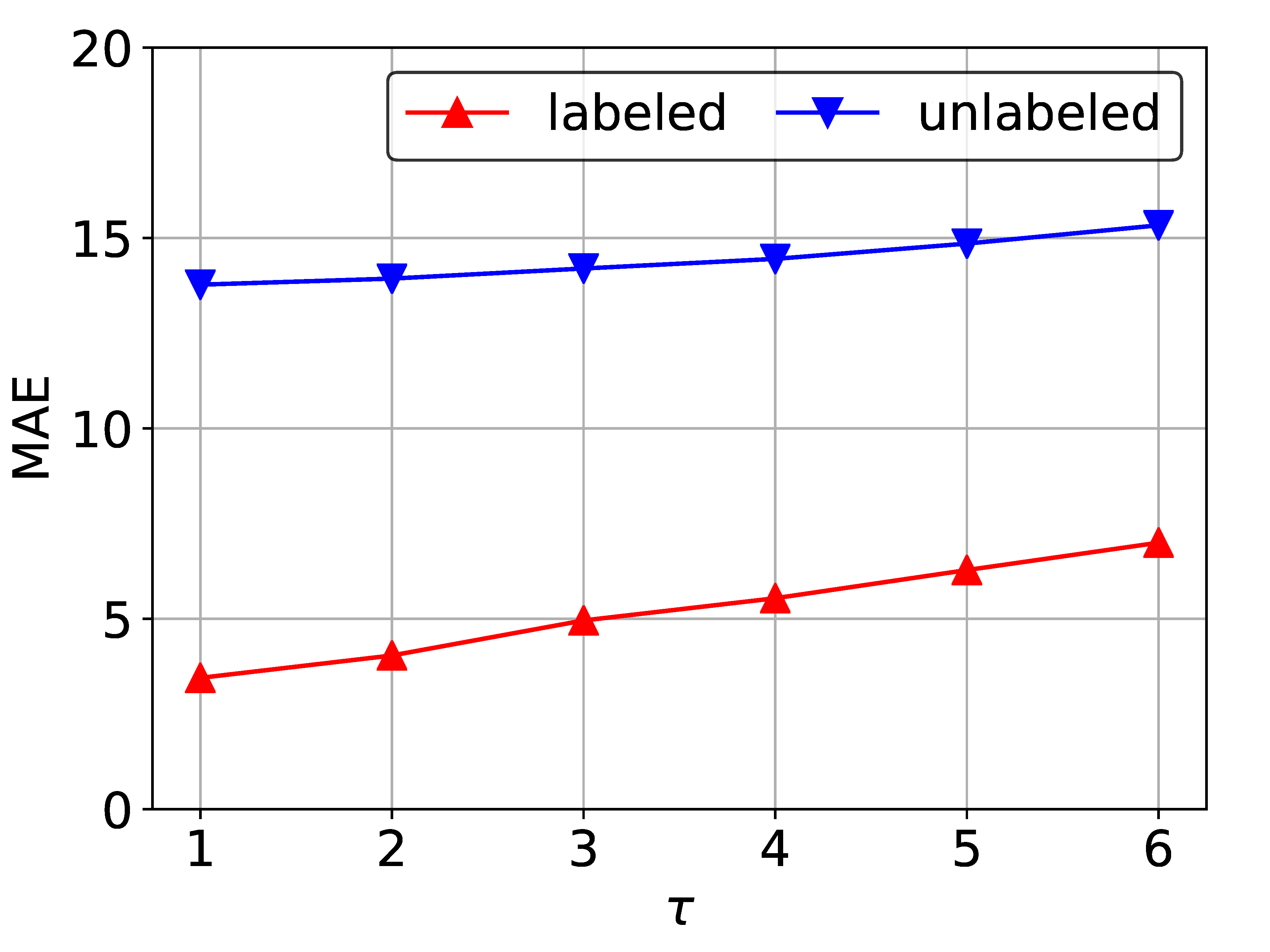}}
\caption{Parameter sensitivity on \beijing.} 
\label{exp:para}
\end{figure*}

\subsubsection{Baselines.} 
We compare our full approach with the following seven baselines and two variants of \hmgnn:
\begin{itemize}
\item \textbf{LR} uses logistic regression for parking availability prediction. We concatenate previous $T$ steps historical features as the input and predict each parking lot separately.
\item \textbf{GBRT} is a variant of boosting tree for regression tasks. It is widely used in practice and performs well in many data mining challenges. We use the version in XGboost~\cite{chen2015xgboost}, and the input is the same as LR.
\item \textbf{GRU}~\cite{chung2014empirical} predicts the PA of each parking lot without considering spatial dependency. We train two GRUs for $P_l$ and $P_u$ separately. 
\item \textbf{Google-Parking}~\cite{arora2019hard} is the parking difficulty prediction model deployed on Google Maps. It uses a feed-forward deep neural network for prediction.
\item \textbf{Du-Parking}~\cite{rong2018parking} is the parking availability estimation model used on Baidu Maps. It fuses several LSTMs to capture various temporal dependencies.
\item \textbf{STGCN}~\cite{yu2018spatio} is a state-of-the-art graph neural network model for traffic forecasting. It models both spatial and temporal dependency with convolution structure. The input graph is constructed as described in the original paper but keeps same graph connectivity with our CxtConv.
\item \textbf{DCRNN}~\cite{li2018dcrnn_traffic} is another graph convolution network based model, which models spatial and temporal dependency by integrating graph convolution and GRU. The input graph is the same as STGCN.
\item \textbf{CxtGNN} is a basic version of \hmgnn, without including PA approximation and soft clustering graph convolution.
\item \textbf{CAGNN} is another variant of \hmgnn but without including the soft clustering graph convolution block.
\end{itemize}

\subsection{Overall performance}
\tabref{table:overall} reports the overall results of our methods and all the compared baselines on two datasets with respect to MAE and RMSE.
As can be seen, our model together with its variants outperform all other baselines using both metrics, which demonstrates the advance of \hmgnn.
Specifically, \hmgnn achieves $(31.8\%, 31.3\%, 29.9\%)$ and $(30.9\%, 31.5\%, 30.9\%)$ improvements beyond the state-of-the-art approach~(DCRNN) on MAE and RMSE on \beijing for $(15min, 30min, 45min)$ prediction, respectively. 
Similarity, the improvement of \emph{MAE} and \emph{RMSE} on \shenzhen are $(29.6\%, 28.7\%, 30.5\%)$ and $(28.8\%, 28.8\%, 28.4\%)$.
Moreover, we observe significant improvement by comparing \hmgnn with its variants~(\ie CxtGNN and CAGNN).
For example, by adding the PA approximation module, CAGNN achieves $(18.6\%, 18.6\%, 18.3\%)$ lower MAE and $(12.2\%, 12.6\%, 13.5\%)$ lower RMSE than CxtGNN on \beijing, respectively.
By further adding the SCConv block, \hmgnn achieves $(14.2\%, 14.1\%, 13.4\%)$ lower MAE and $(20\%, 19.7\%, 18.2\%)$ lower RMSE than CAGNN on \beijing.
The improvement in \shenzhen are consistent.
All above results demonstrate effectiveness of the PA approximation and the hierarchical graph convolution architecture.

Looking further in to the results, we observe all graph convolution based models~(\ie STGCN, DCRNN and \hmgnn) outperform other deep learning based approaches~(\ie Google-Parking and Du-parking), which consistently reveals the advantage of incorporating spatial dependency for parking availability prediction.
Remarkably, GBRT outperforms Google-parking, GRU, LR, and achieves a similar result with Du-parking, which validates our exception that GBRT is a simple but effective approach for regression tasks.
One extra interesting finding is that both MAE and RMSE of all methods on \shenzhen is relatively smaller than on \beijing. This is possible because the spatial distribution of parking lots is more dense and evenly distributed in \shenzhen; therefore they are easier to predict.

\subsection{Parameter sensitivity}
Due to space limitations, here we report the impact of the ratio of labeled parking lot~(\ie $|P_l|/N$), the proportion of latent nodes in the soft clustering graph convolution with respect to the total number of parking lot~(\ie $K/N$), the input time step $T$ and the prediction time step $\tau$ using MAE on \beijing. Each time we vary a parameter, set others to their default values. The results on \beijing using RMSE and on \shenzhen using both metrics are similar. 

First, we vary the ratio of the labeled parking lot from $0.1$ to $0.9$. The results are reported in \figref{exp:para-lr_bj}. The results are unsurprising: equipping more real-time sensors in parking lots enables us to more accurately predict PA. However, equipping more sensors lead to extra economic cost and may be constrained by policies of each parking lot. Finding the most cost-effective ratio and exploring optimal sensor distribution are important problems in the future study.

Then, we vary the ratio of the latent nodes from $0.01$ to $0.8$. For example, there are $1,965$ parking lots on \beijing, $0.01$ corresponds to $20$ latent nodes. The results are reported in \figref{exp:para-hc-bj}. As can be seen, there is a performance improvement by increasing the ratio of latent node form $0.01$ to $0.1$, but a performance degradation by further increasing the ratio of the latent node from $0.1$ to $0.8$. The reason is that heavily reduce the number of latent nodes reduces the discriminative power of learned latent representation, whereas too many latent nodes reduces the regularization power of learned latent representation.

To test the impact of input length, we vary $T$ from $3$ to $18$. The results are reported in \figref{exp:para-tin-bj}. \hmgnn achieves least errors when $T=12$. 
One possible reason is that an excessively short-term input can not provide sufficient temporal correlated information, whereas too long input introduces more noises for temporal dependency modeling.

Finally, to test the impact of prediction step, we vary $\tau$ from $1$ to $6$. The results are reported in \figref{exp:para-tout-bj}. We separate the result of labeled and unlabeled parking lots separately. Overall, labeled parking lots are much easier to predict. Besides, by increase $\tau$, the error of all parking lots increases consistently. However, we can observe the error of labeled parking lots are increasing faster, this makes sense because the temporal dependency between observed PA and future PA becomes lower when $\tau$ goes large.

\begin{figure}[t]
\centering
\subfigure[{\small MAE}]{\label{exp:effect_area_mae_bj}
\includegraphics[width=0.48\columnwidth]{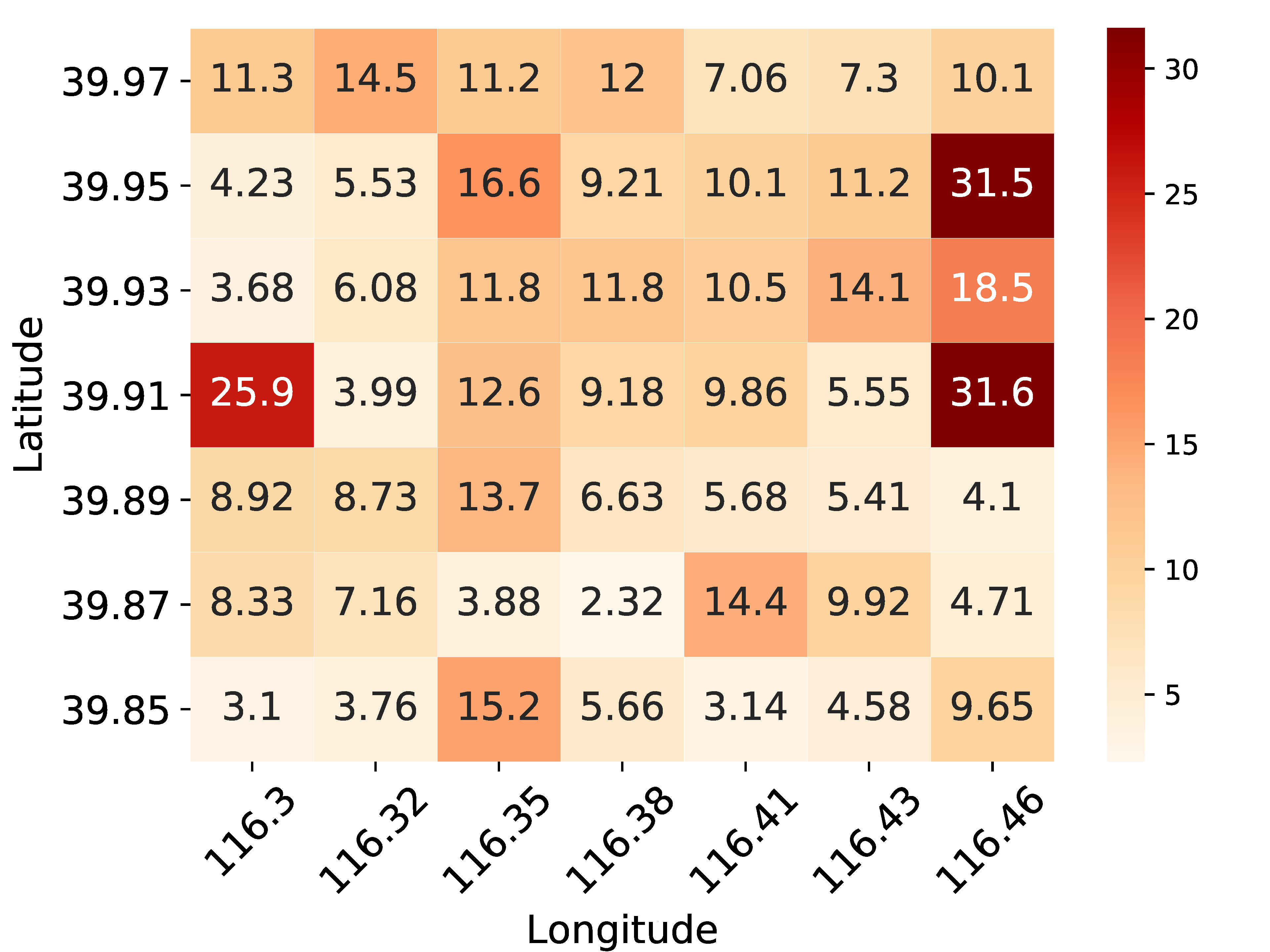}}
\subfigure[{\small $\#$ of parking spot}]{\label{exp:effect_area_parkspace_bj}
\includegraphics[width=0.48\columnwidth]{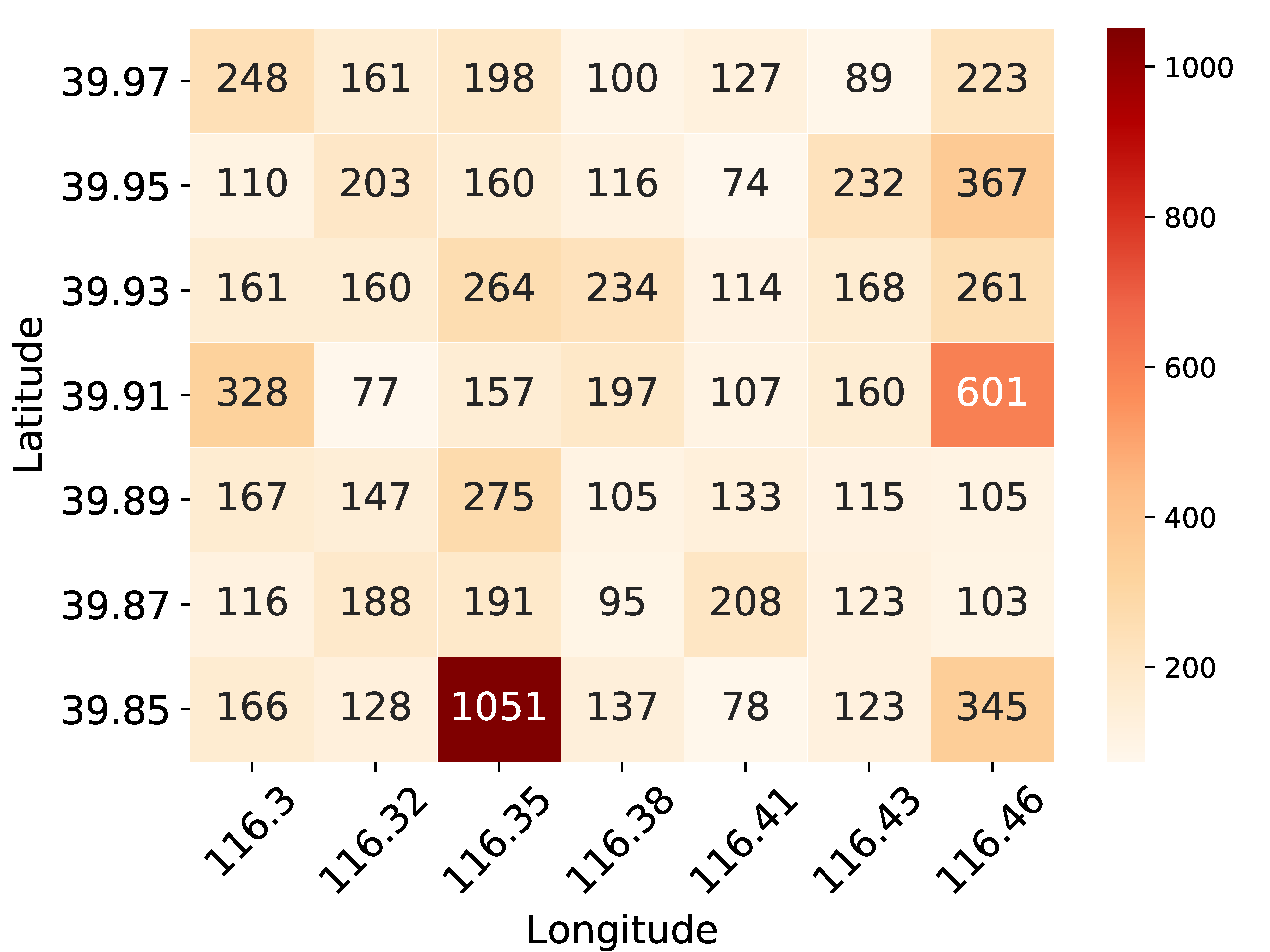}}
\caption{Robustness study on \beijing.} 
\label{exp:effect}
\end{figure}

\subsection{Effectiveness on different regions}
To evaluate the performance of \hmgnn on different regions, we partition \beijing into a set of disjoint grid based on longitude and latitude, and test the performance of \hmgnn on each region. 
\figref{exp:effect_area_mae_bj} and \figref{exp:effect_area_parkspace_bj} plot the averaged MAE of \hmgnn and averaged number of parking spot in each region on \beijing, respectively. 
Overall, the MAE in each region is even except for several outliers.
We find the performance of \hmgnn is highly correlated with the averaged number of parking spots in each region. 
For example, the MAE on region $(116.46, 39.91)$ and $(116.46, 39.95)$ are $31.6$ and $31.5$, which are greater than the overall MAE $10.68$. Meanwhile, the averaged parking spot of these two regions are $601$ and $367$, significantly greater than overall averaged parking spot $210.24$. 
This is possible because for the same ratio of parking availability fluctuate, parking lot with a larger number of parking spot will have larger MAE.
This result indicates in the future further optimization can be applied to these large parking lots to improve the overall performance.

\section{Related Work}\label{sec:related}

\textbf{Parking availability prediction.}
Previous studies on parking availability prediction mainly fall in two categories, contextual data based prediction and real-time sensor based prediction.
For contextual data based prediction, Google-parking~\cite{arora2019hard} and Du-parking~\cite{rong2018parking} predict parking availability based on indirect signals~(\eg user feedbacks and contextual factors), which may induce an inaccurate prediction result. 
For real-time sensor based prediction, study in~\cite{rajabioun2015street} proposes an auto-regressive model 
%for both on-street and off-street parking availability prediction, whereas 
and study in~\cite{fusek2013adaboost} proposes a boosting method
%~\cite{zheng2015parking} proposes both support vector regression and neural network approach 
for parking availability inference.
Above approaches are limited by economic and privacy concerns and are hard to be scaled to all parking lots in a city. Moreover, all the above approaches don't fully exploit non-Euclidean spatial autocorrelations between parking lots, which limits their prediction performance.\\
\textbf{Graph neural network.}
Graph neural network~(GNN) extends the well-known convolution neural network to non-Euclidean graph structures, where the representation of each node is derived by first aggregating and then transforming representations of its neighbors~\cite{velivckovic2017graph}.
It is worth to point out that the idea of our soft clustering graph convolution is partially inspired by~\cite{ying2018hierarchical}, but our objective is to capture global spatial correlation for node-level prediction.
Due to its effectiveness, GNN has been successfully applied to several spatiotemporal forecasting tasks, such as traffic flow forecasting~\cite{li2018dcrnn_traffic,guo2019attention} and taxi demand forecasting~\cite{geng2019spatiotemporal,wang2019origin}. 
However, we argue these approaches either overlook contextual factors or global spatial dependency and are not tailored for parking availability prediction.

\section{Conclusion}\label{sec:conclusion}
In this paper, we present \hmgnn, a city-wide parking availability prediction framework based on both environmental contextual data and partially observed real-time parking availability data. 
We first propose a hierarchical graph convolution module to capture both local and global spatial correlations.
Then, we adopt a simple yet effective GRU module to capture dynamic temporal autocorrelations of each parking lot.
Besides, a parking availability approximation module is proposed for parking lots without real-time parking availability information.
Extensive experimental results on two real-world datasets show that the performance of \hmgnn for parking availability prediction significantly outperforms seven state-of-the-art baselines.
%In the future, we will extend our model for long-term prediction and investigate the influence of improving the coverage ratio of real-time sensors.

\section{Acknowledgement}\label{sec:acknowledgement}
This research is supported in part by grants from the National Natural Science Foundation of China (Grant No.71531001).

\small
\bibliographystyle{aaai} 
\bibliography{6479-aaai}
\end{document}